\documentclass{article}
\pdftrailerid{}
\usepackage[T1]{fontenc}
\usepackage{microtype}
\usepackage{graphicx}
\usepackage{booktabs}
\usepackage{amsmath,amssymb}
\usepackage{xcolor}
\usepackage{float}
\usepackage{array}
\usepackage{hyperref}
\usepackage{tikz}
\usetikzlibrary{arrows.meta,positioning,calc,fit,backgrounds,decorations.pathreplacing}

\usepackage[accepted]{mlsys2025}
\makeatletter
\renewcommand{\mlsys@appearing}{Preprint.}
\renewcommand{\printAffiliationsAndNotice}[1]{%
\stepcounter{@affiliationcounter}%
{\let\thefootnote\relax\footnotetext{\raggedright\hspace*{-\footnotesep}#1%
\forloop{@affilnum}{1}{\value{@affilnum} < \value{@affiliationcounter}}{%
\textsuperscript{\arabic{@affilnum}}\csname @affilname\the@affilnum\endcsname.\ }%
\ifdefined\mlsyscorrespondingauthor@text Correspondence to: \mlsyscorrespondingauthor@text.\fi
\ \\
\Notice@String}}}
\makeatother

\newcommand{\lpsched}{\ell^{\mathrm{sched}}}
\newcommand{\lpdense}{\ell^{\mathrm{dense}}}
\newcommand{\sg}{\operatorname{sg}}
\newcommand{\sysname}{ScaleRLT}

\definecolor{shallowc}{HTML}{1F5FA8}
\definecolor{deepc}{HTML}{D9730D}
\definecolor{gradc}{HTML}{2E8B57}
\definecolor{gatec}{HTML}{8B1E3F}
\mlsystitlerunning{ScaleRLT: Exact Log-Probabilities for RL on Looped Transformers}

\begin{document}

\twocolumn[
\mlsystitle{\sysname: Exact Log-Probabilities and Fast Deferred Rollouts\\for Reinforcement Learning on Looped Transformers}

\begin{mlsysauthorlist}
\mlsysauthor{Dezhen Lu}{tue}
\mlsysauthor{Zixiang Ni}{xjtu}
\mlsysauthor{Runguo Li}{uiuc}
\end{mlsysauthorlist}
\mlsysaffiliation{tue}{University of T\"ubingen, T\"ubingen, Germany}
\mlsysaffiliation{xjtu}{Xi'an Jiaotong University, Xi'an, China}
\mlsysaffiliation{uiuc}{University of Illinois Urbana-Champaign, Urbana, IL, USA}
\mlsyscorrespondingauthor{Dezhen Lu}{dezhen.lu@student.uni-tuebingen.de}
\mlsyscorrespondingauthor{Zixiang Ni}{nzx@stu.xjtu.edu.cn}
\mlsyscorrespondingauthor{Runguo Li}{runguol2@illinois.edu}

\mlsyskeywords{reinforcement learning, looped transformers, recurrent depth, numerical reproducibility, rollout engines}

\vskip 0.3in

\begin{abstract}
Looped language models such as Ouro reuse one block of layers several times per token. A rollout engine can batch the
deeper loops of several response positions and run them only when a deep readout is due. This \emph{deferred} schedule
is fast, but the learner of an on-policy RL algorithm recomputes the sampled tokens' log-probabilities with a dense
pass, and the two computations differ in floating point: in FP32 at 4096 tokens about 18\% of the sampled tokens miss
a $10^{-5}$ same-weight gate, and with bf16 weights nearly every token does. We use a hybrid learner log-probability,
$\ell=\lpdense+\sg(\lpsched-\lpdense)$, whose value is a no-grad replay through the rollout's own schedule and whose
gradient is the dense learner's. The replay reproduces the rollout bit for bit in all 167 FP32 and 28 bf16 real-weight
cells we ran on three platforms, including rollouts whose batch shrinks at stop tokens, and the gradient has the same
FP32 error as other dense implementations in all 102 pre-registered comparisons from 96 to 4096 tokens; the full GRPO
loss of a training step matches a float64 reference to $3.1\times10^{-4}$. On one GPU, deferral raises rollout
throughput by $1.47$--$1.56\times$, and by $1.18$--$1.54\times$ once every step runs from CUDA graphs, bit for bit. In
a two-node GRPO loop, one-wave KV admission and CUDA graphs shorten an update by $2.71\times$ with the gate bitwise at
every update, and four nodes keep the update time within 2\% at the same per-GPU load. In 100-update runs held-out
GSM8K accuracy rose from 23\% to 82\% with the mixed-depth readout that deferral relies on and from 34\% to 90\% with
the native readout, which reached a higher accuracy in less time. For looped models beyond one node we analyse a ring
mapping of layers to GPUs and measure its communication paths.
\end{abstract}
]

\printAffiliationsAndNotice{}
\hypersetup{pdftitle={ScaleRLT: Exact Log-Probabilities and Fast Deferred Rollouts for Reinforcement Learning on Looped Transformers}}

\section{Introduction}
\label{sec:intro}

\begin{figure*}[t]
\centering
\resizebox{\textwidth}{!}{\begin{tikzpicture}[font=\scriptsize, >={Latex[length=1.5mm]}]
  \def\cw{0.56}\def\ch{0.45}
  \tikzset{shal/.style={fill=shallowc!75, draw=shallowc!90!black, line width=0.3pt},
           deep/.style={fill=deepc!25, draw=deepc, line width=0.5pt, rounded corners=1.5pt},
           box/.style={draw, rounded corners=2pt, align=center, inner sep=3pt, line width=0.5pt},
           lab/.style={font=\small\bfseries, anchor=west}}
  \node[lab] at (-0.75, 2.5) {(a) Deferred rollout, $k=4$};
  \foreach \c in {1,...,8} {
    \fill[shal] (\c*\cw-0.5*\cw+0.04, 0.04) rectangle ++(\cw-0.08, \ch-0.08);
    \fill[shal] (\c*\cw-0.5*\cw+0.04, \ch+0.04) rectangle ++(\cw-0.08, \ch-0.08);
    \node[font=\tiny] at (\c*\cw, -0.2) {$t{=}\c$};
  }
  \foreach \s in {0,4} {
    \draw[deep] (\s*\cw+0.5*\cw+0.04, 2*\ch+0.04) rectangle ++(4*\cw-0.08, 2*\ch-0.08);
    \node[font=\tiny, align=center, text=deepc!70!black] at (\s*\cw+2.5*\cw, 3*\ch) {loops 3--4 of $t{=}\the\numexpr\s+1\relax..\the\numexpr\s+4\relax$\\one batch at $t{=}\the\numexpr\s+4\relax$};
  }
  \foreach \c in {1,2,3,5,6,7} { \node[font=\tiny\bfseries, text=shallowc!80!black] at (\c*\cw, 4.4*\ch) {2}; }
  \foreach \c in {4,8} { \node[font=\tiny\bfseries, text=deepc!80!black] at (\c*\cw, 4.4*\ch) {4}; }
  \node[font=\tiny, anchor=east] at (0.42*\cw, 4.4*\ch) {readout depth};
  \foreach \y/\l in {0/1,1/2,2/3,3/4} { \node[font=\tiny, anchor=east] at (0.42*\cw, \y*\ch+0.5*\ch) {loop \l}; }
  \node[anchor=north, align=center, text width=6.0cm] at (4.5*\cw, -0.42) {loops 1--2 run every step (blue); the deeper loops
    wait and run in batches (orange). The function is unchanged, the grouping of every product and reduction is not.};
  \node[lab] at (5.55, 2.5) {(b) Learner log-probability};
  \node[box, fill=shallowc!10, draw=shallowc, minimum width=2.75cm] (roll) at (6.95, 1.85) {rollout log-probs $\ell^{\mathrm{roll}}$};
  \node[box, fill=deepc!12, draw=deepc, minimum width=2.75cm] (rep) at (6.95, 0.8) {scheduled replay, no grad:\\rollout's waves and kernel};
  \node[box, fill=gradc!12, draw=gradc, minimum width=2.75cm] (den) at (6.95, -0.25) {dense learner pass\\(differentiable)};
  \node[box, fill=gatec!10, draw=gatec] (gate) at (10.5, 1.85) {gate: $\lpsched=\ell^{\mathrm{roll}}$ bit for bit,\\else no update};
  \node[box, draw=black!70] (hyb) at (10.5, 0.3) {$\ell=\lpdense+\sg(\lpsched-\lpdense)$\\value $\lpsched$, gradient $\nabla\lpdense$};
  \draw[->] (roll.east) -- (gate.west);
  \draw[->] ($(rep.east)+(0,0.12)$) -- node[above, sloped, font=\tiny, pos=0.5, inner sep=1.5pt] {$\lpsched$} ($(gate.south west)+(0.35,0)$);
  \draw[->] (rep.east) -- ($(hyb.west)+(0,0.12)$);
  \draw[->] (den.east) -- node[below, sloped, font=\tiny, pos=0.5, inner sep=1.5pt] {$\lpdense$} ($(hyb.south west)+(0.45,0)$);
  \node[lab] at (12.55, 2.5) {(c) One GRPO update};
  \node[box, fill=shallowc!10, draw=shallowc, text width=2.0cm] (r) at (13.75, 1.85) {rollout:\\KV waves, graphs};
  \node[box, fill=gatec!10, draw=gatec, text width=2.0cm] (g) at (16.35, 1.85) {gate:\\replay per wave};
  \node[box, fill=gradc!12, draw=gradc, text width=2.0cm] (t) at (16.35, 0.3) {train:\\one optimizer step};
  \node[box, draw=black!70, text width=2.0cm] (p) at (13.75, 0.3) {publish\\new weights};
  \draw[->] (r) -- (g); \draw[->] (g) -- (t); \draw[->] (t) -- (p); \draw[->] (p) -- (r);
  \node[anchor=north, align=center, text width=4.6cm] at (15.05, -0.42) {one optimizer step per rollout batch keeps the
    replay on the rollout's weights};
\end{tikzpicture}}
\caption{\sysname{} at a glance (schematic). (a)~The deferred rollout runs loops 1--2 of every response input at once
and batches loops 3--4 until a deep readout (every $k$-th input) needs them. (b)~The learner's log-probability takes its
value from a no-grad replay through the rollout's own schedule, which the gate requires to equal the rollout bit for bit,
and its gradient from the dense learner (\autoref{eq:t3}). (c)~One GRPO update; one optimizer step per rollout batch keeps
the replay on the rollout's weights.}
\label{fig:overview}
\end{figure*}
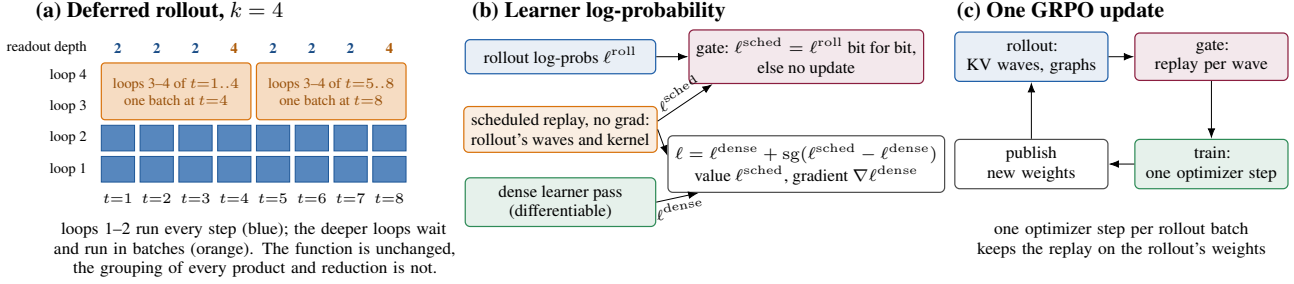

Looped (recurrent-depth) transformers apply the same block of layers several times per token, trading parameters for
depth \citep{universal,huginn,ouro}. Reinforcement learning with verifiable rewards is a natural way to train their
reasoning, and its cost is dominated by rollouts. A looped model makes rollouts unusually expensive: every loop keeps its
own KV history, so both compute and KV memory grow with the number of loops. Ouro runs all four of its loops for every
token. We study a mixed-depth readout policy for it, in which most tokens are read out after two loops and every fourth
token after all four. The deeper loops of the shallow tokens are still needed by later deep tokens, but they can be
computed later and in batches (\autoref{fig:overview}a). This deferred schedule saves rollout time, although, as we find,
the mixed readout it relies on can cost quality. It also changes the grouping of every matrix multiplication and
attention reduction, while the learner of an on-policy algorithm such as PPO or GRPO \citep{ppo,grpo} recomputes the same
log-probabilities with a dense pass. This paper makes the two sides agree exactly, then makes the deferred system fast on
one node and plans it for models that do not fit on one.

\textbf{Contributions.}
\vspace{-0.6\baselineskip}
\begin{itemize}\setlength{\itemsep}{1pt}\setlength{\parskip}{0pt}
  \item \emph{Dense replay is not exact.} In FP32 about 18\% of the sampled tokens miss a $10^{-5}$ same-weight gate at
        4096 tokens, independently of batching, and nearly all do in bf16 (\autoref{sec:dense}).
  \item \emph{An exact learner log-probability.} A no-grad replay through the rollout's own schedule supplies the value
        and the dense learner the gradient (\autoref{sec:t3}). The replay is bitwise in every cell on three platforms, in
        FP32 and bf16, and the gradient is as accurate as any dense gradient, also for the full training loss
        (\autoref{sec:exact}, \autoref{sec:grad}).
  \item \emph{A fast deferred rollout and training loop.} Deferral, one host synchronization per step and per-step CUDA
        graphs speed up rollouts bit for bit; in a two-node GRPO loop one-wave admission and graphs shorten an update by
        $2.71\times$, the loop scales to sixteen GPUs, and we measure what the mixed readout costs in quality
        (\autoref{sec:system}).
  \item \emph{A plan for models beyond one node.} A ring mapping of layers to GPUs keeps loop-shared weights and per-loop
        KV in place; point-to-point measurements set its cost (\autoref{sec:scale}).
\end{itemize}

Every system and learning run had its predictions written down before it was submitted; supplementary
Appendix~\ref{app:prereg} lists every outcome, including the predictions that failed.

\section{Setting}
\label{sec:setting}

\textbf{Model and readout policy.} We use Ouro-1.4B \citep{ouro}: 24 decoder layers applied in four loops, hidden size
2048, 16 attention heads. Every loop keeps its own exact KV history. The first response token is read out after all four
loops of the prompt prefill. Each later response input runs loops 1--2 immediately and is read out at depth~2, except
every $k$-th input ($k=4$ unless noted), which is read out at depth~4. The deferred engine holds the loop-2 states of
pending inputs and runs loops 3--4 for all of them at once when a deep readout is due. An eager variant runs loops 3--4
at every step; it produces the same readouts with a different grouping. $k=1$ reads out all four loops at every token
(``K4''). The mixed-depth readout is our policy, not Ouro's native decoding, which runs all four loops for every token.

\textbf{What deferral changes.} At loop $L$, a token attends only to the loop-$L$ keys and values of earlier positions,
and the input to loop~3 is the token's own loop-2 output. Deferring loops 3--4 therefore does not change the function,
only how the work is grouped: prompt prefill, per-step shallow passes and deferred deep batches put different rows into
each matrix multiplication and split each attention reduction differently from a dense pass. Every causal ancestor of a
deep readout must itself run all four loops, so almost every token runs them eventually; deferral saves time by
batching, not by skipping work.

\textbf{Admission waves.} The production adapter splits a rollout batch into waves under a KV byte budget and a request
cap, computed from the prompt lengths alone. Each wave runs to completion before the next starts; requests never join a
running cohort and leave it when they sample a stop token or reach the token limit.

\textbf{Learner.} The learner is a Megatron \citep{megatron} module that wraps the same native layers, replays the exact
per-loop KV history of each sequence with PyTorch scaled-dot-product attention, and recomputes each layer in the backward
pass. It accumulates gradients in FP32 and feeds a fused log-softmax with a hand-written backward; no Transformer Engine
kernel is involved. The training loop applies one optimizer step per rollout batch.

\section{Dense Replay Does Not Reproduce the Rollout}
\label{sec:dense}

\autoref{tab:dense} compares the rollout's log-probabilities with an independent dense per-sequence pass under the same
FP32 weights (arm A0). The gap grows with length: at 96 tokens about 1\% of the sampled tokens differ by more than
$10^{-5}$, at 4096 tokens about 18\% do, on both platforms and even at batch size one. A $10^{-5}$ same-weight gate is
therefore unpassable at production length, independently of batching. The training loop's own learner exceeded
$10^{-5}$ in 33 of 36 cells at 96 tokens (largest $5.9\times10^{-5}$) and differed from the rollout by up to
$3.5\times10^{-4}$ in the long cells. In FP32 the largest ratio deviation $|\exp(\Delta)-1|$ is below $10^{-3}$, far
inside the PPO clip range, so the problem is the gate, not the size of the update. With bf16 weights the gap is three
orders of magnitude larger: 89--98\% of the sampled tokens miss and the largest difference is 0.45 at 96 tokens and 1.6
at 4096. A bf16 learner inside a training loop (\autoref{sec:learning}) missed by a median 0.16 per response (largest
2.7), yet only 0.02\% of its tokens left the PPO clip range: at 512 tokens the mismatch fails the gate without changing
the update much.

\begin{table}[t]
\centering
\caption{Dense per-sequence replay (A0) against the rollout's log-probabilities, FP32 unless marked bf16. ``Tokens over
$10^{-5}$'' counts sampled tokens whose absolute log-probability difference exceeds the gate.}
\label{tab:dense}
\footnotesize\setlength{\tabcolsep}{3pt}
\resizebox{\columnwidth}{!}{\begin{tabular}{lrrrr}
\toprule
Platform & Length & Cells & Max $|\Delta\log p|$ & Tokens over $10^{-5}$ \\
\midrule
L40S & 96 & 54 & $5.1\times10^{-5}$ & 161/14,496 (1.1\%) \\
L40S & 1024 & 3 & $1.4\times10^{-4}$ & 404/10,016 (4.0\%) \\
L40S & 4096 & 1 & $2.6\times10^{-4}$ & 760/4,096 (18.6\%) \\
GH200 & 96 & 36 & $2.6\times10^{-5}$ & 65/9,664 (0.7\%) \\
GH200 & 1024 & 4 & $8.9\times10^{-5}$ & 599/17,984 (3.3\%) \\
GH200 & 4096 & 5 & $8.0\times10^{-4}$ & 13,545/73,408 (18.5\%) \\
\midrule
GH200, bf16 & 96 & 20 & $4.5\times10^{-1}$ & 5,260/5,920 (88.9\%) \\
GH200, bf16 & 1024 & 2 & $6.7\times10^{-1}$ & 8,552/8,992 (95.1\%) \\
GH200, bf16 & 4096 & 1 & $1.6$ & 3,997/4,096 (97.6\%) \\
\bottomrule
\end{tabular}
}
\end{table}

\section{A Hybrid Log-Probability Takes Its Value from the Rollout's Schedule}
\label{sec:t3}

Let $\lpsched_t$ be the log-probability of sampled token $t$ from a no-grad, teacher-forced replay through the rollout's
own schedule, and $\lpdense_t$ the dense learner's. The learner uses
\begin{equation}
  \ell_t \;=\; \lpdense_t + \sg\!\left(\lpsched_t - \lpdense_t\right),
  \label{eq:t3}
\end{equation}
where $\sg$ stops the gradient. Its value is $\lpsched_t$ and its gradient is $\nabla\lpdense_t$
(\autoref{fig:overview}b). If the scheduled replay is bitwise equal to the rollout, the importance ratio at
$\theta_{\mathrm{old}}$ is exactly one and the strict gate can be kept. The correction is exact only while the weights
equal those of the rollout, which holds because the loop takes one optimizer step per rollout batch; with several steps
per batch, $\lpsched$ would have to be recomputed at each step. The replay needs no gradient, so it can use the rollout's
own attention kernel; a fully differentiable scheduled learner would need a functional KV cache, step-level
checkpointing and a differentiable attention backend, and we use one only as a reference arm (A3) at short lengths.

\textbf{Arms.} A0: dense per sequence. A1: the batch packed into one dense pass with block-diagonal causal attention.
A2: no-grad replay through the rollout's schedule on the rollout's backend. A2x: the same schedule on the other attention
backend (Triton $\leftrightarrow$ PyTorch). A2-solo: the same backend, each request replayed alone. A3: the rollout's
schedule executed differentiably. M: the production learner of \autoref{sec:setting}.

\section{Scheduled Replay Is Bitwise Exact}
\label{sec:exact}

\textbf{Setup.} FP32 weights unless marked bf16, temperature 1, full vocabulary, eight GSM8K-style \citep{gsm8k}
prompts, one sampling seed per request. Grid cells vary the batch size (1/2/4/8), the readout schedule (K4, eager,
deferred), the length (96, 1024, 4096 tokens), the backend and the seed; request $i$ of a cell stops after $L-8i$
tokens, so batch membership changes during the cohort. The production-rollout cells instead call the production
generation routine with stop tokens and a 1024-token limit. Each hardware platform is a separate cohort with the same
pinned set of 137 Python packages (PyTorch 2.12.1 with CUDA 13.0, Triton 3.7.1).

\textbf{Results.} \autoref{tab:replay} lists every real-weight cell. A2 is bitwise in all 167 of them. The
production-rollout cells include 36 whose batch shrank at stop tokens (166 stopped and 20 length-capped responses), so
the replay also covers membership changes that the engine, not the test, decides. Triton and PyTorch rollouts sampled
identical tokens in every pair we compared, yet A2x is never bitwise, and neither is A2-solo for $B\ge2$. Exactness
therefore requires the rollout's batch composition, step schedule and kernel together; in production this means
replaying each admission wave with the rollout's partition and backend, which the adapter records.

\textbf{bf16.} A subset of the grid and the production-rollout cells with bf16 weights, using the same arm code: 28
cells, 34{,}732 sampled tokens, lengths 96 to 4096, batch 1 to 8, both backends, two seeds, and five production cells
whose batches shrank at stop tokens. A2 is bitwise in all 28; A2x is never bitwise, with differences up to 1.2, and
A2-solo is never bitwise for $B\ge2$. The exactness of the scheduled replay is a property of the schedule and kernel,
not of FP32.

\begin{table*}[t]
\centering
\caption{Scheduled replay (A2) against the rollout, FP32 above the line and bf16 below it. ``A2 bitwise'' counts cells
in which every sampled token's log-probability is reproduced exactly. A2x and A2-solo give the largest absolute
difference over all cells. The CPU arms ran PyTorch attention only, so they have no A2x. Lengths are token limits;
production-rollout responses end earlier at stop tokens.}
\label{tab:replay}
\small\setlength{\tabcolsep}{4pt}
\begin{tabular}{lrrllrrr}
\toprule
Platform & Cells & Tokens & Lengths & $B$ & A2 bitwise & A2x max & A2-solo max \\
\midrule
L40S (CPU arms) & 10 & 2,928 & 96/512 & 1/2/4/8 & 10/10 & -- & $2.2\times10^{-5}$ \\
L40S & 58 & 28,608 & 96/1024/4096 & 1/2/4/8 & 58/58 & $2.1\times10^{-5}$ & $2.1\times10^{-5}$ \\
GH200 & 45 & 101,056 & 96/1024/4096 & 1/2/4/8 & 45/45 & $9.5\times10^{-5}$ & $1.0\times10^{-4}$ \\
GH200, production rollout & 54 & 55,678 & $\le$1024 & 1/2/4/8 & 54/54 & $4.0\times10^{-5}$ & -- \\
\midrule
GH200, bf16 & 23 & 19,008 & 96/1024/4096 & 1/2/4/8 & 23/23 & $1.2$ & $4.4\times10^{-1}$ \\
GH200, bf16, production rollout & 5 & 15,724 & $\le$1024 & 8 & 5/5 & $4.4\times10^{-1}$ & -- \\
\bottomrule
\end{tabular}

\end{table*}

\section{The Hybrid Gradient Is as Accurate as Any Dense Gradient}
\label{sec:grad}

\textbf{Truth and criterion.} For each cell we compute the gradient of $\sum_i w_i \sum_t \ell_{i,t}$ with two
weightings, alternating-sign advantages and all ones. The reference is a float64 evaluation of the same function with
float64 normalization, softmax and log-softmax and the engine's rotary constants, accumulated per sequence. Two
independent float64 implementations, the learner itself and a functional dense executor, agree to $4.7\times10^{-14}$ in
relative gradient norm on real weights. We report $e(X)=\|\nabla_X-\nabla_{\mathrm{truth}}\|/\|\nabla_{\mathrm{truth}}\|$
and fixed the criterion before running: $e(\mathrm{M}) \le 2\max_Y e(Y)$, where $Y$ ranges over other legitimate dense
implementations (A0 and A1 at 96 tokens; A0 and the learner on the math attention backend at longer lengths).
Long-sequence rollouts disable stop tokens so that every response reaches its limit, the one deviation from production
settings; development seeds are reported apart from fresh ones.

\textbf{Results.} \autoref{tab:grad} summarizes the 102 comparisons; the criterion holds in all of them. The production
learner's relative error grows from about $10^{-5}$ at 96 tokens to about $10^{-4}$ at 1024 tokens and stays there at
4096. The single $1.3\times10^{-3}$ outlier is an ill-conditioned K4 cell whose gradient norm is less than half that of
the others; every implementation is poor there. The criterion is lenient by design: at 96 tokens the explicit-softmax
executor was up to $14\times$ further from truth than the learner in an ill-conditioned seed, while the learner was never
more than $1.6\times$ worse than either alternative; at 512--4096 tokens the dense executor was up to $1.9\times$ closer
in some cells and the math-backend learner up to $3\times$ further in others. Neither is consistently more accurate. The
gradient half of \autoref{eq:t3} adds nothing beyond the FP32 noise that any dense learner already has.

\textbf{The full training loss.} A further pre-registered check took the gradient that the real training step
accumulates just before its optimizer step: GRPO advantages normalized within each group, the T3 value of
\autoref{eq:t3} inside the PPO objective, the mean over each sample's response tokens, and accumulation over 32 or 64
micro-batches with the step's normalization. One update of 64 samples was recomputed by an independently written float64
reference, with float64 normalization and the advantages rebuilt from the raw rewards. The relative error of the FP32
gradient was $3.1\times10^{-4}$ with the native readout and $5.7\times10^{-6}$ with the mixed one (cosine similarities
0.99999996 and 0.99999999998), and the clipped gradient norm the training stack logged matched the reference norm to
$1.3\times10^{-4}$. Three deliberately wrong losses (a token mean over the whole batch, advantages without the
standard-deviation division, and full-depth log-probabilities in place of the recorded readouts, which differs only for
the mixed readout) were 2{,}500 to $4\times10^{6}$ times further away, so the check would have caught a defect in the
loss path. The native readout's larger relative error comes with an eight times smaller gradient norm for that batch.

\begin{table*}[t]
\centering
\caption{Gradient error of the production learner (M) against float64 truth. ``Other dense, median'' pools the
alternative dense implementations of the same cells. The last column is the largest ratio of the learner's error to the
worse alternative in the same cell; the pre-registered criterion requires it to be at most 2.}
\label{tab:grad}
\small\setlength{\tabcolsep}{4pt}
\begin{tabular}{lrrrrrr}
\toprule
Platform & Length & Comparisons & $e(\mathrm{M})$ median & $e(\mathrm{M})$ max & Other dense, median & $e(\mathrm{M})/\max e(Y)$, max \\
\midrule
GH200 & 96 & 72 & $9.9\times10^{-6}$ & $6.7\times10^{-5}$ & $8.7\times10^{-6}$ & 1.62 \\
B300 & 512 & 2 & $7.2\times10^{-5}$ & $7.2\times10^{-5}$ & $5.6\times10^{-5}$ & 1.09 \\
B300 & 1024 & 20 & $1.2\times10^{-4}$ & $1.3\times10^{-3}$ & $2.0\times10^{-4}$ & 1.19 \\
B300 & 4096 & 8 & $1.1\times10^{-4}$ & $1.5\times10^{-4}$ & $9.5\times10^{-5}$ & 0.96 \\
\bottomrule
\end{tabular}

\end{table*}

\section{Deferred Rollouts Are Fast Inside the Training Loop}
\label{sec:system}

All measurements here are FP32 on GH200 GPUs, with stop tokens disabled for single-GPU throughput and enabled in the
training loop.

\subsection{One GPU}
\label{sec:onegpu}

\textbf{Deferral.} In eight comparisons timed on the same node against the eager schedule, which reads out at the same
depths but runs loops 3--4 at every step, deferral raised throughput by $1.47$--$1.56\times$, from batch 1 to 64 and from
512 to 4096 tokens, up to 79\,GiB, with identical peak memory (Table~\ref{tab:deferral}). K4 runs at the
eager rate, so the saving comes from batching loops 3--4, not from doing less work. Scheduled replay took
$0.86$--$0.96\times$ the rollout time at the rollout's peak memory and was bitwise in all three configurations measured.

\textbf{One synchronization per step and CUDA graphs.} A profile of the deferred engine showed the GPU busy for only
26--36\% of each step: sampling read each row's token back to the host one row at a time, and a step issued 3{,}000 to
4{,}500 CUDA operations. We draw every row on the device and read all tokens and log-probabilities back once per step,
keeping each request's random-number stream and per-row operations unchanged, and replay each per-step pass of 24 layers
from a CUDA graph captured per row count and shared across loop depths. The synchronization change alone gives
$1.01$--$1.20\times$; with graphs the rollout is $1.59$--$2.49\times$ faster than before, most at small batch
(Table~\ref{tab:graphs}). Rollouts are bitwise equal to the per-row loop with or without graphs, and the
scheduled replay from graphs is $1.44$--$1.72\times$ faster and bitwise in every pairing of graph and eager rollout and
replay, so the replay of \autoref{eq:t3} may use graphs whether or not the rollout did.

\textbf{Deferral with CUDA graphs.} Graphs remove much of the launch overhead that deferral amortizes, so we measured
deferral again with graphs on (\autoref{tab:deferralgraphs}); each run builds a fresh runner that captures its own graphs,
as the production engine does. Deferral is still $1.18$--$1.54\times$ faster than the eager schedule, most at small
batch, where the GPU is latency-bound; per token, its forward time is $0.62$--$0.79\times$ the eager one. Host time is
5--10\% of a step at batch 8 and about 30\% at batch 64, so the forward pass, not sampling, remains the main cost. With
stop tokens the batches keep changing shape, so deferral captures about twice as many graphs (41 against 21), and capture
takes 27\% of its rollout; without capture time its advantage would be $1.37\times$. Keeping the KV pool and its graphs
across the admission waves of one update recovers almost none of it: in a pre-registered check the second wave's 6--14
new graphs were nearly all row counts the first wave never saw, sharing saved at most one, and the rollout was bitwise
unchanged. The captures that repeat are those of each update's first wave and the gate's replay of the rollout's row
counts; avoiding them needs the pool to outlive the rollout, which on one GPU competes with training for memory.

\begin{table*}[t]
\centering
\caption{One GH200, FP32, CUDA graphs (tokens per second). ``Eager'' reads out at the same depths as ``Deferred''; K4
reads every token at depth 4. ``Def./eager'' is the deferred/eager ratio, ``W/o capture'' the same ratio without graph-capture time, and ``Capture''
that time's share of the deferred rollout. One timing per cell.}
\label{tab:deferralgraphs}
\small\setlength{\tabcolsep}{4pt}
\begin{tabular}{lrrrrrr}
\toprule
Config & Deferred & Eager & K4 & Def./eager & W/o capture & Capture \\
\midrule
$B=8$, 512 & 316 & 205 & 213 & 1.54 & 1.56 & 2\% \\
$B=64$, 512 & 1211 & 1025 & 976 & 1.18 & 1.19 & 1\% \\
$B=24$, 2048 & 591 & 452 & 442 & 1.31 & 1.31 & 1\% \\
$B=32$, 512, stop tokens & 485 & 412 & 416 & 1.18 & 1.37 & 27\% \\
\bottomrule
\end{tabular}

\end{table*}

\subsection{Two Nodes, Eight GPUs}
\label{sec:dp8}

\textbf{Setup.} Two nodes with four GH200 GPUs each run one data-parallel GRPO job of eight ranks
(\autoref{fig:overview}c). Each update samples four responses for each of eight prompts per rank (32 per rank, 256 in
total) with stop tokens and a 512-token limit, gates on the scheduled replay, takes one optimizer step and publishes the
new weights to the rollout engine. The gradient reduce-scatter and all-gather of 5.6\,GB across the two nodes took
0.12\,s in a calibration run, against about 15\,s of training per update. The same calibration showed that raising the
wave cap from 8 to 16 requests cut rollout and gate time to $0.60\times$, which led to the one-wave policy below. We
report the median of updates 2 and 3, each at the slowest rank.

\textbf{Results.} \autoref{tab:dp8} lists five configurations on the same software. Admitting each rank's 32 samples in
one wave, instead of two waves of 16, cut rollout and gate time to $0.46\times$ and $0.45\times$; replaying the per-step
passes from CUDA graphs in both cut them by a further $0.51\times$ and $0.46\times$. Together the update went from
195.6\,s to 70.5\,s ($2.77\times$), or to 72.2\,s ($2.71\times$) with the memory release described next. Training is now
the largest phase, 43\% of the update. The gate was bitwise at every rank and update (120 of 120), and the training stack
reported a PPO KL estimate and a learner--rollout log-probability difference of exactly zero at every update. At the
first update the graph rollouts equal the eager rollouts bit for bit on all eight ranks, and a separate job on other
nodes reproduces them bit for bit. The one-wave and two-wave rollouts sampled the same tokens, but their
log-probabilities differ by up to $6.0\times10^{-4}$: wave composition changes the floating-point result, which is why
the gate replays each wave with the rollout's own partition.

\textbf{Memory.} One-wave admission sizes a wave by an estimate of KV bytes, and the estimate is accurate (allocated
memory grew by at most $1.012\times$ the estimate), yet the GPUs filled to 97{,}208 of 97{,}871\,MiB and in one
configuration an NCCL buffer allocation failed during the final checkpoint save. Releasing cached allocator memory before
each rollout and before training (``release'' in \autoref{tab:dp8}) lowered the peak to 63{,}016\,MiB at the same speed.
Allocator snapshots then traced a remaining 14\,GiB of pinned memory to one 109\,MiB block kept alive by a loop variable;
freeing it removed the defect (Appendix~\ref{app:memory}).

\begin{table*}[t]
\centering
\caption{Two nodes $\times$ four GH200, data-parallel GRPO, 32 samples per rank, 512-token limit, FP32. Seconds are the
median of updates 2--3 at the slowest rank. ``Gate bitwise'' counts rank-updates whose scheduled replay reproduced the
rollout exactly. Peak memory is the largest per-GPU value reported by \texttt{nvidia-smi} (out of 97{,}871\,MiB) before
any checkpoint save. ``Release'' returns cached allocator memory to the device before each rollout and before training.}
\label{tab:dp8}
\small\setlength{\tabcolsep}{3pt}
\begin{tabular}{lrrrrrrrr}
\toprule
Configuration & Waves & Rollout s & Gate s & Train s & Step s & Rollout tok/s & Gate bitwise & Peak MiB \\
\midrule
Cap 16 (2 waves) & 2 & 80.1 & 77.6 & 32.5 & 195.6 & 1,336 & 24/24 & 67,062 \\
One wave & 1 & 36.5 & 34.7 & 29.9 & 106.2 & 2,929 & 24/24 & 97,208 \\
One wave + graphs & 1 & 18.7 & 15.9 & 30.6 & 70.5 & 5,728 & 24/24 & 97,034 \\
One wave, release & 1 & 37.0 & 34.2 & 30.3 & 106.9 & 2,892 & 24/24 & 63,016 \\
One wave + graphs, release & 1 & 19.2 & 16.1 & 31.5 & 72.2 & 5,578 & 24/24 & 63,246 \\
\bottomrule
\end{tabular}

\end{table*}

\subsection{Four Nodes, Sixteen GPUs}
\label{sec:dp16}

A pre-registered job ran the final two-node configuration on four nodes with four arms (\autoref{tab:dp16}). With the
per-rank load unchanged (512 samples per update), the update took 81.2\,s with the native readout and 69.2\,s with the
mixed one, against 79.6 and 69.4\,s on eight GPUs: weak scaling within 2\%, the gate bitwise at all 640 rank-updates,
peak memory unchanged and no pinned memory. Learning matched the two-node runs: 84.2\% and 76.4\% after 20 updates
(84.6\% and 75.8\% on two nodes), with the native readout again ahead on 68 problems against 29 (exact McNemar
$p=9\times10^{-5}$). Strong scaling did not meet its prediction: with 16 samples per rank, training halved, from 32.3 to
15.9\,s, but the rollout (21.6 against 22.4\,s) and the gate (20.1 against 20.1\,s) did not move, so the update fell to
$0.76\times$ rather than the predicted $0.65\times$. Both phases are bounded by the longest response, 512 decode steps of
one graph replay each, which a smaller wave does not shorten; adding ranks compresses only training, and the update
cannot fall below about 42\,s in this configuration. Turning activation recomputation off cut training from 32.0 to
19.1\,s ($0.60\times$) with the first gradient norm bit for bit identical and the peak unchanged.

\begin{table*}[t]
\centering
\caption{Four nodes $\times$ four GH200, data-parallel GRPO, FP32, one job. Seconds are the median of updates 2 onward
at the slowest rank; ``weak'' keeps 32 samples per rank (512 per update), ``strong'' keeps the two-node total of 256
(16 per rank). Accuracy is held-out GSM8K before training and after the last update.}
\label{tab:dp16}
\small\setlength{\tabcolsep}{3pt}
\begin{tabular}{lrrrrrrrrr}
\toprule
Arm & Samples & Updates & Update (s) & Rollout & Gate & Train & Peak (MiB) & Gate bitwise & Accuracy \\
\midrule
K4, weak & 512 & 20 & 81.2 & 23.4 & 20.9 & 32.0 & 49,582 & 320/320 & 33.8\% $\to$ 84.2\% \\
Mixed, weak & 512 & 20 & 69.2 & 18.5 & 16.2 & 30.3 & 50,000 & 320/320 & 23.0\% $\to$ 76.4\% \\
K4, strong & 256 & 10 & 60.2 & 21.6 & 20.1 & 15.9 & 39,262 & 160/160 & 33.8\% $\to$ 79.6\% \\
K4, no recompute & 512 & 5 & 68.3 & 23.7 & 21.0 & 19.1 & 49,518 & 80/80 & -- \\
\bottomrule
\end{tabular}

\end{table*}

\subsection{Learning}
\label{sec:learning}

\textbf{Setup.} We trained each readout policy, the mixed readout of \autoref{sec:setting} and the native K4 readout, for
100 GRPO updates. The pre-registered runs use the two-node setup: four responses for each of 64 GSM8K training prompts per
update (256 in total, 32 per rank in one wave), stop tokens, a 512-token limit, temperature 1 and learning rate
$10^{-6}$. A reduced version on one GH200 with 16 prompts per update ran first. Data order and seeds are the same across
runs. Before training and every 20 updates, each run scores one sample on each of the first 500 GSM8K test problems with
the training reward (a correct answer inside \verb|\boxed{}|), reusing the same per-problem seeds.

\textbf{Results.} \autoref{tab:learning} shows held-out accuracy. On two nodes the mixed readout improved by 59.4 points
(300 problems gained, 3 lost) and K4 by 56.2 (288 gained, 7 lost). Part of the gain is learning to stop within 512 tokens
(the stopped fraction rose from 41\% and 49\% to 98\% and 97\%), but not all: among responses that stopped, accuracy rose
from 25\% to 84\% and from 34\% to 91\%. We did not keep the response texts, so we cannot separate learning to box the
answer from better reasoning. The gate was bitwise at every rank and update (1{,}600 of 1{,}600 on two nodes, 200 of 200
on one GPU), with a PPO KL estimate of exactly zero; this is also the first evidence of the hybrid learner at batch for
K4. Initial accuracies are identical across setups, as the initial weights and evaluation seeds are.

\begin{table}[t]
\centering
\caption{Held-out accuracy on 500 GSM8K test problems (one sample each, same seeds at every evaluation) during GRPO with
the mixed readout and K4, FP32. The two-node runs are pre-registered; the one-GPU runs use a quarter of the samples per
update. One training seed per run.}
\label{tab:learning}
\footnotesize\setlength{\tabcolsep}{3pt}
\begin{tabular}{rrrrr}
\toprule
Updates & \multicolumn{2}{c}{8 GPUs, 256/update} & \multicolumn{2}{c}{1 GPU, 64/update} \\
 & Mixed & K4 & Mixed & K4 \\
\midrule
0 & 23.0\% & 33.8\% & 23.0\% & 33.8\% \\
20 & 75.8\% & 84.6\% & 69.2\% & 82.6\% \\
40 & 78.8\% & 87.8\% & 74.4\% & 84.4\% \\
60 & 81.6\% & 87.6\% & 80.8\% & 86.6\% \\
80 & 82.0\% & 89.0\% & 81.0\% & 87.6\% \\
100 & 82.4\% & 90.0\% & 78.8\% & 87.8\% \\
\bottomrule
\end{tabular}

\end{table}

\textbf{The mixed readout costs quality.} K4 was 10.8 points ahead before training and still 7.6 ahead after 100 updates
on two nodes (9.0 on one GPU). On the last evaluation K4 alone solved 52 problems that the mixed readout missed, against
14 the other way (exact McNemar $p=2.8\times10^{-6}$; one training seed per policy). The mixed readout's update took
69.4\,s against 79.6\,s ($0.87\times$), with the rollout $1.23\times$ faster per token. K4 reached 84.6\% after 20
updates, about 27 minutes, already above the mixed readout's 82.4\% after 100 updates, about 1.9 hours. In this
configuration the mixed readout is not worth its speed. On one GPU the gap was smaller ($0.96\times$): training took 57\%
of the update, and two waves per update each captured their own graphs, so the rollout was only $1.09\times$ faster per
token.

\textbf{Single-GPU training was host-bound, and is no longer.} With the same software, the single GPU was busy a median
33--34\% of the sampled time against 79--92\% on the cluster nodes, bf16 training took as long as FP32, and the autograd
thread held one core at its maximum clock. Two pre-registered checks removed the host cost. Keeping the activations
instead of recomputing them cut the training phase from 116.6 to 62.9\,s per update ($0.54\times$) with the rollout
bitwise identical, the first gradient norm within $8\times10^{-8}$ and the peak unchanged, because the peak is set by the
rollout's KV budget. Running every sequence of a micro-batch through the learner at once, right-padded to the longest
and one slab per loop over the prefixes the exact-KV schedule requires (float64 agreement with the per-sequence path to
$10^{-10}$ in a CPU test), cut it to 35.4\,s with two sequences per micro-batch ($0.30\times$ the original), 0.55\,s per
sample against 1.01\,s on the two-node run, with the GPU busy 92--98\% of the time and the rollout bitwise identical
(Table~\ref{tab:batched}). The batched gradient differs from the per-sequence one by $4\times10^{-4}$ in
norm; against float64 both FP32 paths were $1$--$7\times10^{-4}$ from the truth, the batched one closer on the K4
schedule and farther on a mixed one, so they are of the same accuracy class rather than equivalent.

\textbf{bf16, and training without the gate.} A pre-registered follow-up repeated the reduced mixed-readout run with a
bf16 learner and rollout engine (FP32 master weights and optimizer state) in two arms: the hybrid learner with its gate,
and training on the dense log-probabilities with no gate. The gate stayed bitwise at all 100 updates in bf16 with a PPO
KL estimate of exactly zero. The ungated arm saw the mismatch at every update, yet reached the same accuracy, 77.0\%
against 77.2\% (\autoref{tab:bf16train}; 42 problems solved by the ungated arm alone and 43 by the gated one, exact
McNemar $p=1$, and no evaluation during training separated them, $p\ge0.21$). Only 0.02\% of its tokens left the PPO clip
range, so at 512 tokens the deviation is neither caught by the clip nor large enough to hurt; at this scale the hybrid
learner buys certainty, not accuracy. Both bf16 arms trailed the FP32 run for the first 40 updates (51\% against 69\%
after 20) and ended 1.6--1.8 points behind; a consistent but untested explanation is the checkpoint's bf16 weight grid,
which lets at most 26\% of the weights leave their initial value within 20 Adam updates at a learning rate of $10^{-6}$.
The bf16 update was $0.89\times$ the FP32 one (201.8 against 227.3\,s), not the predicted $0.8\times$; training itself
took as long, as expected of a host-bound step.

\begin{table}[t]
\centering
\caption{Held-out accuracy during GRPO with the mixed readout on one GH200, bf16 learner and rollout engine, 16 prompts
per update: the hybrid learner with its gate (T3) and the dense learner with no gate. The FP32 column is the reduced run
of \autoref{tab:learning}. One training seed per run.}
\label{tab:bf16train}
\footnotesize\setlength{\tabcolsep}{4pt}
\begin{tabular}{rrrr}
\toprule
Updates & bf16, T3 & bf16, dense ungated & FP32, T3 \\
\midrule
0 & 23.6\% & 23.6\% & 23.0\% \\
20 & 50.8\% & 50.0\% & 69.2\% \\
40 & 69.2\% & 69.0\% & 74.4\% \\
60 & 72.4\% & 70.0\% & 80.8\% \\
80 & 74.0\% & 72.6\% & 81.0\% \\
100 & 77.2\% & 77.0\% & 78.8\% \\
\bottomrule
\end{tabular}

\end{table}

\section{Looped Models Map onto a GPU Ring}
\label{sec:scale}

This section is a design with measurements of its communication paths; the mapped engine is not yet implemented.

\textbf{When one node is not enough.} With a distributed optimizer over $D$ data-parallel replicas, each replica holds
about $6+12/D$ bytes per parameter. For three looped shapes at $D=8$ the rollout weights alone fit one node of four
95\,GiB GPUs, but not the combination that RL on one set of GPUs requires: learner state, a second copy of the weights
for the rollout engine, and KV for every loop. For a 69B shape the learner state and rollout weights are already
613\,GiB (Table~\ref{tab:scaling}).

\textbf{The mapping.} Place the layers on $S$ GPUs of $N$ nodes as contiguous arcs, one arc per node, connected in a
ring. Each loop sends the hidden state once around the ring: $S$ boundary crossings, of which only the $N$ node
boundaries use the network; $N$ is a lower bound for any placement on $N$ nodes. On nodes of four GPUs three quarters of
the crossings stay on NVLink. A layer's $K$ loops run on the same GPU, so the gradient of the shared weights accumulates
locally and data-parallel traffic stays at one copy of the parameters; the KV of each layer and loop never moves; and a
GPU's layers are visited $K$ times per forward pass, like the $K$ virtual stages of an interleaved pipeline schedule
\citep{interleaved}, whose bubble shrinks by about $K$ (existing schedules assume distinct layers per virtual stage).
Deferral does not reduce the bytes that cross the ring, but it groups loops 3--4 into fewer and larger messages.

\textbf{Against sharded data parallelism.} FSDP and ZeRO-3 \citep{fsdp,zero} gather each layer's weights when the layer
runs. A looped model runs each layer $K$ times forward and $K$ times backward per micro-batch, so it regathers every
weight $2K$ times unless it keeps the gathered weights and gives up the memory saving. The ring moves no weights: for an
8192-token micro-batch, FSDP gathers 324--1032 times more bytes per GPU than one ring boundary carries.

\textbf{Measured communication paths.} \autoref{tab:p2p} reports point-to-point bandwidth between two nodes of four
GH200 GPUs, connected by NVLink inside a node and by one 200\,Gb/s NIC per GPU with GPUDirect RDMA between nodes. A NIC
carries 23.8\,GB/s, 95\% of its line rate, four pairs run concurrently at nearly that rate each, and the remote GPU
index does not matter on this switched fabric. NCCL send/recv over NVLink reached only 51\,GB/s inside an eight-rank
communicator, because NCCL assigns point-to-point channels per peer from the minimum over all peers and network peers
get two by default; a node-local communicator or eight channels per network peer give 124\,GB/s without slowing the
NIC, and CUDA IPC with the copy engine gives 132\,GB/s. Splitting a node boundary over four NICs with an NVLink scatter
and gather delivers 92\,GB/s at 256\,MiB in eight pipelined chunks; at 64\,MiB and below chunking does not pay. A
host-staged transfer reached 4.9\,GB/s, so activations should never leave the GPU, while a 1\,KiB host message took
38\,$\mu$s one way. Small messages are bound by the host cost of each call, about 60\,$\mu$s in
\texttt{torch.distributed}; an eight-hop ring of 64\,KiB messages took 159\,$\mu$s per loop, and 203\,$\mu$s when captured
in a CUDA graph. With these bandwidths a per-hop model, which matched the measured 256\,MiB rings within 1\%, puts the
forward activations of one loop around a two-node ring for the 69.3B shape at 17.8\,ms per micro-batch with tuned NVLink
channels and 9.4\,ms with the four-NIC split, against 26.9\,ms with NCCL defaults, while the forward and backward compute
per stage and loop is about 1050\,ms under an assumed 400\,TFLOP/s, so the ring's communication can be hidden. The open
costs are the pipeline bubble, memory balance across arcs, and the per-call host overhead of the many small messages of
decoding.

\begin{table}[t]
\centering
\caption{Point-to-point bandwidth between two nodes of four GH200 GPUs, unidirectional GB/s, NCCL send/recv unless
noted, median of three repeats. ``8 channels per net peer'' sets NCCL's per-network-peer channel count to 8. The boundary
split uses the best of 1, 4, 8 or 16 pipelined chunks at each size.}
\label{tab:p2p}
\footnotesize\setlength{\tabcolsep}{3pt}
\resizebox{\columnwidth}{!}{\begin{tabular}{lrrr}
\toprule
Path & 4 MiB & 64 MiB & 256 MiB \\
\midrule
NIC, one pair & 21.8 & 23.8 & 23.8 \\
NIC, one pair to a different GPU index & 21.5 & 23.6 & 23.7 \\
NVLink, 8-rank communicator, defaults & 45.6 & 51.4 & 51.4 \\
NVLink, node-local communicator & 73.3 & 121.4 & 124.3 \\
NVLink, 8-rank, 8 channels per net peer & 68.7 & 122.1 & 124.3 \\
NVLink, copy engine via CUDA IPC & 90.7 & 129.3 & 132.0 \\
Node boundary split over 4 NICs, 8 channels per net peer & 13.4 & 90.4 & 92.6 \\
\bottomrule
\end{tabular}
}
\end{table}

\section{Related Work}
\label{sec:related}

Looped and recurrent-depth transformers \citep{universal,huginn,ouro} reuse layers to add depth at fixed parameter
count; we apply a mixed-depth readout policy to Ouro and make its rollouts exact and fast. RL frameworks for language
models such as HybridFlow \citep{hybridflow} co-locate or separate rollout and training engines, and RL objectives such
as PPO and GRPO \citep{ppo,grpo} rely on the learner reproducing the rollout's probabilities. When separate rollout and
training stacks disagree, the update is effectively off-policy; truncated importance sampling corrects for the mismatch
with capped learner-to-rollout probability ratios \citep{tis}, and training in FP16 instead of bf16 reduces it
\citep{fp16rl}. Our hybrid learner removes the mismatch from the value instead of correcting for it, and keeps the dense
learner only for the gradient. Paged KV management \citep{pagedattention} underlies our admission budget. Pipeline and
sharded data parallelism \citep{megatron,interleaved,zero,fsdp} are the baselines for \autoref{sec:scale}; the loop
structure changes their trade-off because layers are reused within one forward pass.

\section{Limitations}
\label{sec:remaining}

\textbf{Learning quality.} The learning evidence is one seed per readout policy and setup, 100 updates of 256 (two nodes)
or 64 (one GPU) samples, FP32, and a 500-problem evaluation with one sample per problem.
\textbf{Precision and optimizer.} The scheduled replay is exact in FP32 and bf16, and the gate stayed bitwise through 100
bf16 training updates, but the gradient and throughput results are FP32, and the bf16 training evidence is one seed on one
GPU at 512 tokens, where the ungated learner did as well. Whether the gate matters for learning at longer sequences, where
the dense mismatch is ten times larger, is open. The runs used Megatron's PyTorch Adam fallback.
\textbf{Gradient criterion.} The criterion of \autoref{sec:grad} is on a surrogate loss; the full GRPO loss was checked
against float64 on one batch at 512 tokens. Under T3 the ratio is exactly one, so the PPO clipping branch never runs.
\textbf{Readout policy.} Deferral needs the mixed-depth readout, which trailed the native one by 7.6--9 points after 100
updates while an update was only 4--13\% shorter. We tried one readout interval ($k=4$).
\textbf{Scale.} One model (Ouro-1.4B), up to 4096 tokens in the exactness study and up to four nodes of data parallelism,
where strong scaling is bounded by the per-token latency of rollout and gate. The mapping of \autoref{sec:scale} is a
design backed by point-to-point measurements.

\section{Conclusion}
\label{sec:conclusion}

A deferred rollout of a looped transformer and a dense learner compute the same function in different floating-point
orders, so a strict same-weight gate fails at production length. Replaying the rollout's own schedule restores bitwise
agreement in every cell we ran, in FP32 and bf16, while the dense learner keeps supplying a gradient as accurate as any
dense gradient. With that gate in place, deferral, one host synchronization per step, CUDA graphs and one-wave admission
make the training loop fast and keep it bitwise on two and four nodes. The mixed readout that deferral needs, however,
learned less than the native readout in our runs, which reached a higher accuracy in less time.

\bibliography{refs}
\bibliographystyle{mlsys2025}

\clearpage
\onecolumn
\appendix
\section{Single-GPU Rollout Tables}
\label{app:onegpu}

\begin{table}[H]
\centering
\caption{Single GH200, FP32, stop tokens disabled: eager and deferred schedules with identical readouts (tokens per
second, median of two timings). Peak memory is the same for both. Two cells give no clean ratio: eager and deferred ran
on different nodes in one, and in the other the two deferred timings differ by 10\% and the ratio is far below its
neighbours for a reason we did not find.}
\label{tab:deferral}
\small\setlength{\tabcolsep}{4pt}
\begin{tabular}{rrrrrrl}
\toprule
Length & $B$ & Eager tok/s & Deferred tok/s & Ratio & Peak GiB & Note \\
\midrule
512 & 1 & 10.8 & 16.5 & 1.53 & 6.2 &  \\
512 & 8 & 85.2 & 128.7 & 1.51 & 12.0 &  \\
512 & 32 & 269.1 & 402.8 & 1.50 & 31.9 &  \\
512 & 64 & 479.7 & 705.2 & 1.47 & 58.4 &  \\
2048 & 8 & 83.0 & 127.3 & 1.53 & 30.0 &  \\
2048 & 16 & 139.8 & 236.1 & -- & 54.6 & different nodes \\
2048 & 24 & 260.3 & 392.4 & 1.51 & 79.2 &  \\
4096 & 4 & 56.2 & 85.9 & 1.53 & 29.7 &  \\
4096 & 8 & 105.5 & 132.9 & 1.26 & 54.0 & unexplained \\
4096 & 12 & 140.2 & 218.3 & 1.56 & 78.3 &  \\
\bottomrule
\end{tabular}

\end{table}

\begin{table}[H]
\centering
\caption{Single GH200, FP32, deferred schedule, stop tokens disabled (tokens per second). ``Per-row sync'' is the engine
before the change. ``Bitwise'' means the one-sync and the graph rollouts both reproduce the per-row rollout exactly. Peak
memory is unchanged.}
\label{tab:graphs}
\small\setlength{\tabcolsep}{4pt}
\begin{tabular}{rrrrrrc}
\toprule
$B$ & Length & Per-row sync & One sync/step & + CUDA graphs & Speedup & Bitwise \\
\midrule
8 & 512 & 138 & 165 & 342 & 2.49$\times$ & yes \\
64 & 512 & 888 & 894 & 1412 & 1.59$\times$ & yes \\
24 & 2048 & 336 & 396 & 654 & 1.95$\times$ & yes \\
12 & 4096 & 194 & 213 & 308 & 1.59$\times$ & yes \\
\bottomrule
\end{tabular}

\end{table}

\section{The Pinned-Memory Defect}
\label{app:memory}

One-wave admission sizes a wave by an estimate of KV bytes, and the estimate is accurate: the allocated memory grew
during the rollout by at most $1.012\times$ the estimate. The GPUs nevertheless filled up, to 97{,}208\,MiB of
97{,}871\,MiB, and in one configuration an NCCL buffer allocation failed during the final checkpoint save, which then
hung until the collective timed out. Releasing cached allocator memory before each rollout and before training lowered
the peak to 63{,}016\,MiB at the same speed. Allocator logging then showed one remaining defect: once per rank, about
14\,GiB stayed reserved after the release. Allocator snapshots on one GPU traced it to a single 109\,MiB block, the FP32
logits of the last sequence of a diagnostic dense replay that runs between the rollout and the gate. A loop variable
kept the block alive; it was allocated in a KV segment the rollout had just freed, so neither the gate nor the release
could reuse that segment. Freeing the variable removed the defect: in the training runs of
Section~\ref{sec:learning}, the reserved memory after the release equalled that before the rollout at every one of 200
updates. The KV budget of one-wave admission is now derived from free device memory only, so memory pinned in this way
cannot be counted as available.

\section{Batched Learner Forward}
\label{app:batched}

\begin{table}[H]
\centering
\caption{Batched learner forward on one GH200 (K4, FP32, 64 samples per update, 3 updates per arm; mean of updates
2--3): $N$ sequences per learner micro-batch, the per-sequence or batched path, and activation recomputation off or on.
Train peak is the allocator's peak during the training phase; the first update's gradient norm shows the rounding
difference between the paths. The rollout was bitwise identical in every arm. Without recomputation the activations of
a padded slab grow by about 12\,GiB per sequence, so three sequences are the most that fit next to the rollout's
80\,GB peak.}
\label{tab:batched}
\small
\begin{tabular}{rllrrrrr}
\toprule
$N$ & Path & Recompute & Train (s) & per sample (s) & GPU busy & Train peak (GiB) & Gradient norm \\
\midrule
1 & per sequence & off & 62.8 & 0.98 & 84\% & 44.8 & 3.44061 \\
1 & batched & off & 54.1 & 0.85 & 73\% & 40.4 & 3.44061 \\
2 & batched & off & 35.4 & 0.55 & 92\% & 51.2 & 3.43930 \\
3 & batched & off & 36.2 & 0.57 & 96\% & 64.4 & 3.43887 \\
4 & batched & off & \multicolumn{5}{l}{out of memory} \\
8 & batched & off & \multicolumn{5}{l}{out of memory} \\
4 & batched & on & 40.2 & 0.63 & 97\% & 33.5 & 3.43869 \\
8 & batched & on & 39.1 & 0.61 & 97\% & 35.6 & 3.43918 \\
16 & batched & on & 36.3 & 0.57 & 97\% & 39.8 & 3.43918 \\
32 & batched & on & 36.6 & 0.57 & 98\% & 49.0 & 3.43918 \\
\bottomrule
\end{tabular}

\end{table}

\section{Memory Arithmetic for the Ring Mapping}
\label{app:scaling}

\begin{table}[H]
\centering
\caption{Three looped shapes with $K=4$ loops, GQA with 8 KV heads, SwiGLU width $3.5h$. Learner state per replica
assumes a distributed optimizer over $D=8$ replicas. The last three columns are per 8192-token micro-batch, forward and
backward, in bf16: the weight bytes FSDP gathers per GPU, the activation bytes one ring boundary carries (all $K$ loops),
and their ratio. Model arithmetic, not measurements.}
\label{tab:scaling}
\small\setlength{\tabcolsep}{4pt}
\begin{tabular}{lrrrrrr}
\toprule
Shape & Params & States/replica & Rollout weights & FSDP regather & Ring boundary & Ratio \\
\midrule
$h=4096$, 48 layers & 10.9B & 76 GiB & 20 GiB & 174 GB & 0.54 GB & 324 \\
$h=6144$, 60 layers & 29.7B & 207 GiB & 55 GiB & 475 GB & 0.81 GB & 590 \\
$h=8192$, 80 layers & 69.3B & 484 GiB & 129 GiB & 1108 GB & 1.07 GB & 1032 \\
\bottomrule
\end{tabular}

\end{table}

\section{Pre-Registered Predictions and Their Outcomes}
\label{app:prereg}

Each run below had its predictions written down before it was submitted. We list every prediction that failed. The
single-GPU CUDA-graph measurement of Section~\ref{sec:onegpu} was not pre-registered. Three further submissions failed
before measuring anything and were repeated after the cause was fixed: a launcher missing from the path, a network
plugin that fell back to sockets because of a missing C++ runtime symbol, and a hang caused by creating a process
subgroup that not all ranks joined in the same way.

\begin{table}[H]
\centering
\caption{Outcomes of the pre-registered system, gradient, communication and learning runs.}
\label{tab:prereg}
\footnotesize\setlength{\tabcolsep}{3pt}
\begin{tabular}{>{\raggedright\arraybackslash}p{4.6cm}rr>{\raggedright\arraybackslash}p{9.2cm}}
\toprule
Run & Predictions & Met & Not met (predicted; measured) \\
\midrule
Two-node calibration, 16 samples/rank & 6 & 5 & Update time on 8 GPUs at most $1.05\times$ that on 4; $1.066\times$ \\
Two-node admission and CUDA graphs & 9 & 8 & Peak memory at most 90\,GB; 97{,}208\,MiB \\
Two-node memory rerun & 8 & 7 & Reserved memory before training within 2\,GiB of that before the rollout; 13.8--14.2\,GiB
  once per rank \\
Allocator snapshots, one GPU & 4 & 4 & -- \\
Point-to-point, first round & 10 & 6 & NCCL NVLink at least 90\,GB/s; 52. NVLink over NIC $4$--$10\times$; $2.2\times$.
  Four-NIC split (unpipelined) at most $0.5\times$ the direct time; $0.54\times$. 4\,KiB half round trip at most
  20\,$\mu$s (NVLink) and 40\,$\mu$s (NIC); 101 and 89\,$\mu$s \\
Point-to-point, second round & 8 & 5 & Single-process copy-engine slowness is host-side; it is not. Host time per call
  10--60\,$\mu$s; 61. CUDA-graph ring at most 15\,$\mu$s per hop; 25 \\
bf16 replay check & 3 & 3 & -- \\
Deferral with CUDA graphs & 5 & 4 & Host time at least 40\% of a step at batch $\le12$; 5--10\%. (A first attempt with
  deterministic kernels was voided: it slowed every schedule several-fold, unlike production) \\
Four nodes, sixteen GPUs (Section~\ref{sec:dp16}) & 7 & 6 & Strong-scaling update at most $0.65\times$ the eight-GPU one;
  $0.76\times$ (rollout and gate are per-token latency) \\
Learning, two nodes (Section~\ref{sec:learning}) & 6 & 5 & Mixed-readout update at most $0.8\times$ the K4 update; $0.87\times$ \\
Learning, one GPU (Section~\ref{sec:learning}) & 6 & 5 & The same; $0.96\times$ (training was 57\% of an update, and the
  rollout only $1.09\times$ faster per token) \\
bf16 training, one GPU (Section~\ref{sec:learning}) & 7 & 5 & Ungated arm's clipped fraction at least 1\%;
  0.02\%. bf16 gated update at most $0.8\times$ the FP32 one; $0.89\times$ (training itself took as long) \\
No recompute, one GPU (Section~\ref{sec:learning}) & 5 & 5 & -- \\
Full GRPO-loss gradient (Section~\ref{sec:grad}) & 5 & 3 & Batched FP32 gradient within $2\times$ the per-sequence
  error; $2.04\times$ (native readout). Every wrong loss at least $10\times$ further; the full-depth one equals the
  native readout's correct loss (design error) \\
Cross-wave KV pool (Section~\ref{sec:onegpu}) & 5 & 3 & Later-wave captures at most half; 18 of 20 and 10 of 10.
  Rollout and gate at most $0.9\times$; $1.08\times$ and $0.98\times$ \\
Batched forward (Section~\ref{sec:learning}) & 9 & 3 & Four sequences fit without recomputation; out of memory.
  Gradient norm within $10^{-4}$ of the per-sequence path; $4\times10^{-4}$. Dense metric at most $0.6\times$; $0.8\times$
  (already compute-bound). Batched FP32 gradient within $2\times$ the per-sequence error against float64 on both
  schedules; one of two \\
\bottomrule
\end{tabular}
\end{table}

\end{document}